\documentclass{article} % For LaTeX2e
\usepackage{nips13submit_e,times}
\usepackage{hyperref}
\usepackage{url}
\usepackage{graphicx}
\usepackage{amsmath,amssymb} % define this before the line numbering.
\usepackage{tabularx}
\usepackage{caption}
\usepackage{subcaption}
\usepackage{array}

\usepackage{times}
\usepackage{epsfig}
\usepackage{graphicx}
\usepackage{amsmath}
\usepackage{amssymb}
\usepackage{url}

\usepackage{color}
\usepackage{colortbl}
\usepackage{caption}
\usepackage{subcaption}
\usepackage{tabularx}
\usepackage{array}
\usepackage{multirow}
\usepackage{float}

\title{Recurrent Fully Convolutional Networks for Video Segmentation}

\newcommand\blfootnote[1]{%
  \begingroup
  \renewcommand\thefootnote{}\footnote{#1}%
  \addtocounter{footnote}{-1}%
 \endgroup
}

\begin{document}

\maketitle
\blfootnote{* Authors contributed equally}
%%%%%%%%% ABSTRACT
\begin{abstract}
Image segmentation is an important step in most visual tasks. While convolutional neural networks have shown to perform well on single image segmentation, to our knowledge, no study has been done on leveraging recurrent gated architectures for video segmentation. Accordingly, we propose and implement a novel method for online segmentation of video sequences that incorporates temporal data. The network is built from a fully convolutional network and a recurrent unit that works on a sliding window over the temporal data. We use convolutional gated recurrent unit that preserves the spatial information and reduces the parameters learned. Our method has the advantage that it can work in an online fashion instead of operating over the whole input batch of video frames. The network is tested on video segmentation benchmarks in Segtrack V2 and Davis. It proved to have 5\% improvement in Segtrack and 3\% improvement in Davis in F-measure over a plain fully convolutional network.
\end{abstract}

%%%%%%%%% BODY TEXT
\section{Introduction}
The recent trend in convolutional neural networks has dramatically changed the landscape in computer vision. The first task that was improved with this trend was object recognition\cite{krizhevsky2012imagenet}\cite{simonyan2014very}\cite{szegedy2015going}. An even harder task that greatly progressed is semantic segmentation, which provides per pixel labelling as introduced in \cite{long2015fully} \cite{zheng2015conditional}\cite{visin2015reseg}. In \cite{long2015fully} fully convolutional network was introduced. These networks yield a coarse segmentation map for any given image, and it is followed by upsampling within the network to get dense predictions. This method enabled an end-to-end training for the task of semantic segmentation of images. However, one missing element in this recent trend is that real-world is not a set of static images. Large portion of the information that we infer from the environment comes from motion. For example, in an activity recognition task, the difference between walking and standing is only profound if you consider a sequence of images.

\begin{figure}[ht!]
     \centering
    \includegraphics[width=0.9\textwidth]{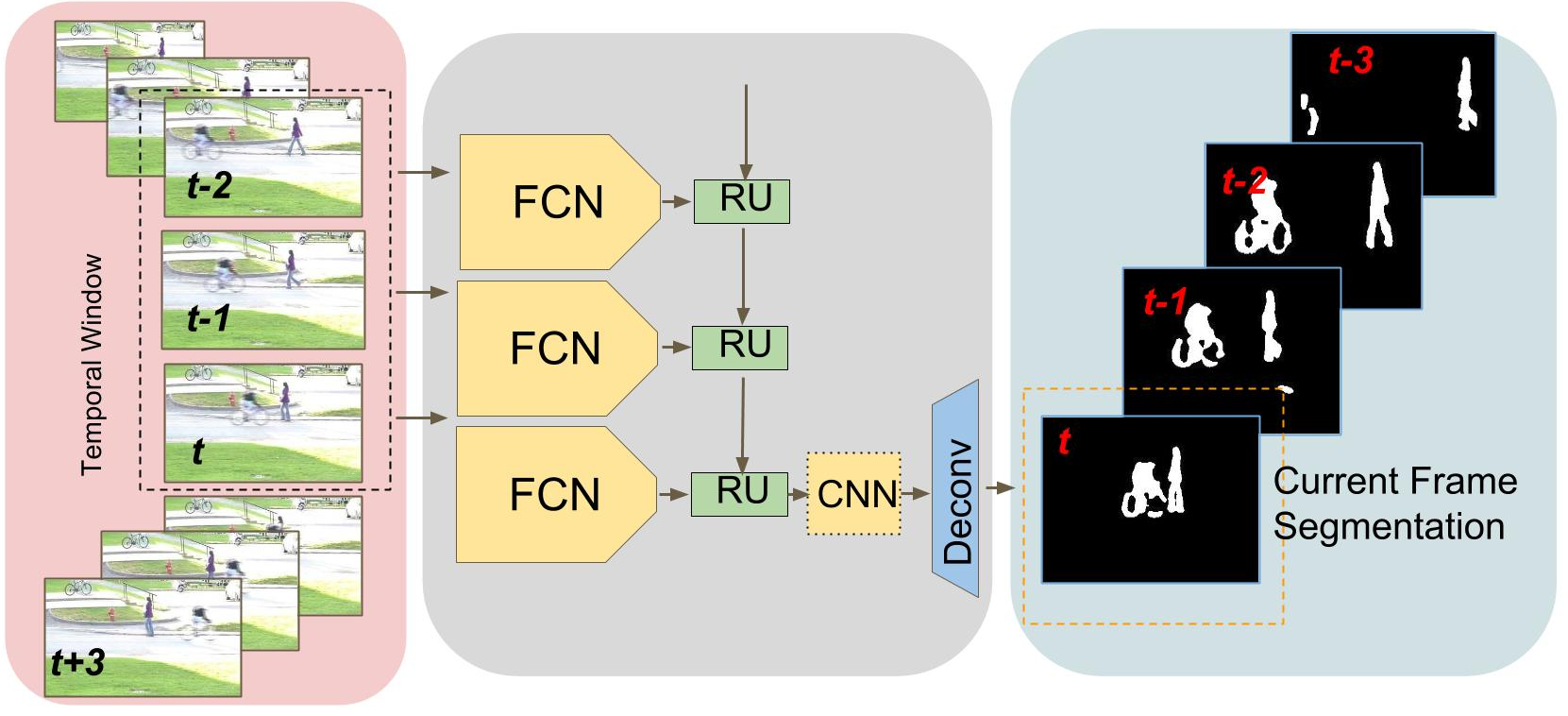}
    \caption{Overview of the Proposed Method of Recurrent FCN. The recurrent part is unrolled for better visualisation}
    \label{fig:overprop}
\end{figure}

The conventional Convolutional Neural Networks (CNN) are not designed to include temporal changes. The simplest way to include temporal information in CNN is to concatenate multiple frames and feed it as a single input. Small variations of this method are used for context classification on one million youtube videos\cite{karpathy2014large}. Surprisingly, it could not improve on single frame prediction by much which can indicate the inefficiency of this approach.  Michalski et al.\cite{michalski2014modeling} created a network which learns transformations between frames of a long video sequence. In \cite{taylor2010convolutional} convolutional restricted Boltzman machine is introduced which learns optical-flow-like features from input image sequence. Another proposed method \cite{pavel2015recurrent} uses Recurrent Neural Networks (RNN) which have shown their power in different tasks. It introduces a combination of CNN and RNN for video segmentation. However, their RNN architecture is sensitive to initialization, and training is difficult due to the vanishing gradient problem. Their design does not allow usage of a pre-trained network and it cannot process large input images as the number of parameters in their network grows exponentially with the input size.

Several architectures are proposed that aim to solve the main bottleneck of recurrent networks namely, vanishing or exploding gradients. In \cite{hochreiter1997long} Long Short Term Memory (LSTM) is presented. It is used for various applications such as generating sequences for text \cite{graves2013generating}, dense image captioning\cite{johnson2015densecap} and video captioning \cite{donahue2015long}. Another recently proposed architecture is the Gated Recurrent Unit (GRU) \cite{cho2014properties}. It was shown in \cite{chung2014empirical} that LSTM and GRU outperform other traditional recurrent architectures, and that GRU showed similar performance to LSTM but with reduced number of parameters. One problem with these previous architectures is that they only work with vectorized sequences as input. Thus, they are incapable of handling data where spatial information is critical like images or feature maps. Another work\cite{ballas2015delving} uses convolutional GRU for learning spatiotemporal features from videos that tackle this issue. In their work, experiments on video captioning and human action recognition were conducted.

The usage of fully convolutional networks (FCN) combined with recurrent gated units can solve many of the pitfalls of the previous approaches. In this paper, we present : (1) A novel architecture that can incorporate temporal data directly into FCN for video segmentation. We have chosen the Recurrent Neural Network as the foundation of our structure since it is shown to be effective in learning temporal dynamics. (2) An end-to-end training method for online video segmentation that does not need to process data offline. Overview of the suggested method is presented in Figure \ref{fig:overprop}, where a sliding window over the frames is used and passed through the recurrent fully convolutional network(RFCN). To our knowledge, this is the first work that presents a recurrent fully convolutional network for pixel-wise labelling.

The paper is structured as follows. In Section \ref{background} we will discuss the preliminary topics, then the proposed method will be presented in details in section \ref{method}. It is followed by the experiments section and discussion of the results in section \ref{experiments}. Finally, section \ref{conclusion} concludes the paper and presents potential future directions.

%-------------------------------------------------------------------------

\section{Background} \label{background}
This section will review FCN and RNN which will be repeatedly referred to through the article.

\subsection{Fully Convolutional Networks (FCN)}
In convolutional neural networks that are used for classification, the few last fully connected layers are responsible for the classification part. But, with pixel-wise labelling, there is a need for dense predictions on all the pixels. In \cite{lin2015efficient} the idea of using a fully convolutional neural network that is trained for pixel-wise semantic segmentation is presented. It is shown that it surpasses the state of the art in semantic segmentation on PASCAL VOC, NUYDv2, and SIFT Flow datasets. The FCN method is briefly discussed in what follows.

FCN architecture is based on VGG\cite{simonyan2014very} architecture due to its success in classification tasks. However, due to the fully connected layers that these networks have, they can only accept fixed size input and produce a classification label. To overcome this problem, it is possible to convert a fully connected layer into a convolutional layer. Accordingly, this network can yield coarse maps pixel wise prediction instead of one classification output. 

In order to have dense prediction from this coarse map, it needs to be up-sampled to the original size of the input image. The up-sampling method can be a simple bi-linear interpolation. But in \cite{lin2015efficient} a new layer that applies upsampling within the network was presented. It makes it efficient to learn the up-sampling weights within the network using back-propagation. The filters of the deconvolution layer act as the basis to reconstruct the input image. Another idea for up-sampling is to stitch together output maps from shifted version of the input. But It was mentioned in \cite{lin2015efficient} that using up-sampling with deconvolution is more effective. In \cite{noh2015learning} the idea of having a full deconvolution network with both deconvolution layers and unpooling layers is presented.

The FCN architecture has been tried in different applications. In \cite{huang2015densebox} it is used for object localization. In \cite{wang2015visual} a modified architecture was used for visual object tracking. Finally for semantic segmentation in \cite{noh2015learning} a full deconvolution network is presented with stacked deconvolution layers.

%======================================================= RNN ======================================================
\subsection{Recurrent Neural Networks}

Recurrent Neural Networks\cite{vinyals2012revisiting} are designed to incorporate sequential information into a neural network framework. These networks are capable of learning complex dynamics by utilizing a hidden unit in each recurrent cell. This unit works like a dynamic memory that can be changed based on the state that the unit is in. Accordingly, the process of each unit yields to two outcomes. Firstly, an output is computed from the current input and the hidden units values (the networks memory). Secondly, the network updates its memory based on, again, current input and hidden units value. 
The simplest recurrent unit can be modeled as\ref{eq:rnn}. 
\begin{subequations}
\label{eq:rnn}
\begin{align}
    h_t &= \theta\phi(h_{t-1}) + \theta_x x_t \\
    y_t &= \theta_y \phi(h_t)
\end{align}    
\end{subequations}
Here, $h$ is the hidden layer, $x$ is the input layer and $y$ is the output layer and $\phi$ is the activation function.

Recurrent networks were successful in many tasks in speech recognition and text understanding\cite{sutskever2011generating} but they come with their challenges. Unrestricted data flow between units causes problems with vanishing and exploding gradients \cite{bengio1994learning}. During the back propagation through recurrent units, the derivative of each node is dependent of all the nodes which processed earlier. This is shown in equations \ref{vanishing_eq1},\ref{vanishing_eq2} and \ref{eq:explode} where $E$ is the loss of the layer. To compute $\frac{\partial{h_t}}{\partial{h_k}}$ a series of multiplication from $k=1$ to $k = t -1$ is required. Assume that $\dot{\phi}$ is bounded by $\alpha$ then $||{\frac{\partial{h_t}}{\partial{h_k}}}|| < \alpha^{t-k}$

\begin{equation} 
    \frac{\partial{E}}{\partial{\theta}} = \sum_{t=1}^{t = S}\frac{\partial{E_t}}{\partial{\theta}}
    \label{vanishing_eq1}
\end{equation}

\begin{equation}
    \frac{\partial{E_t}}{\partial{\theta}} = \sum_{k=1}^{k = t}\frac{\partial{E_t}}{\partial{y_t}}\frac{\partial{y_t}}{\partial{h_t}}\frac{\partial{h_t}}{\partial{h_k}}\frac{\partial{h_k}}{\partial{\theta}}
    \label{vanishing_eq2}
\end{equation}

\begin{equation}
    \frac{\partial{h_t}}{\partial{h_k}} = \prod_{i=k+1}^{t}\frac{\partial{h_i}}{\partial{h_{i-1}}} = \prod_{i=k+1}^{t}\theta^{T}diag[\dot{\phi}(h_{i-1})]
    \label{eq:explode}
\end{equation}

A solution to this problem is to use gated structures. The gates can control back propagation flow between each node. Long-Short Term Memory \cite{hochreiter1997long} is the first such proposed architecture and it is still popular. A more recent architecture is Gated Recurrent Unit \cite{cho2014properties} which has simpler cells yet with competent performance \cite{chung2014empirical}.   

\subsubsection{Long Short Term Memory (LSTM)}
As mentioned, LSTM uses a gated structure where each gate controls the flow of a particular signal. Each LSTM node has three gates that are input, output and forget gate each with learnable weights. These gates can learn the optimal way to remember useful information from previous states and decide the current state. In equations \ref{eq:lstm} the procedure of computing different gates and hidden states is shown, where $i_t$, $f_t$ and $o_t$ are input, forget and output gates respectively. While $c_t$ denote the cell internal state, and $h_t$ is the hidden state. 
\begin{subequations}
\begin{align}
    i_t &= \sigma(W_{xi}x_t + W_{hi}h_{t-1} + b_i) \\
    f_t &= \sigma(W_{xf}x_t + W_{hf}h_{t-1} + b_f) \\ 
    o_t &= \sigma(W_{xo}x_t + W_{ho}h_{t-1} + b_o) \\
    g_t &= \sigma(W_{xc}x_t + W_{hc}h_{t-1} + b_c) \\ 
    c_t &= f_t\odot c_{t-1}+i_t \odot g_t \\ 
    h_t &= o_t \odot \phi(c_t)
\end{align}
\label{eq:lstm}
\end{subequations}

\subsubsection{Gated Recurrent Unit (GRU)}
\begin{figure}[ht]
    \centering
    \includegraphics[width=0.9\linewidth]{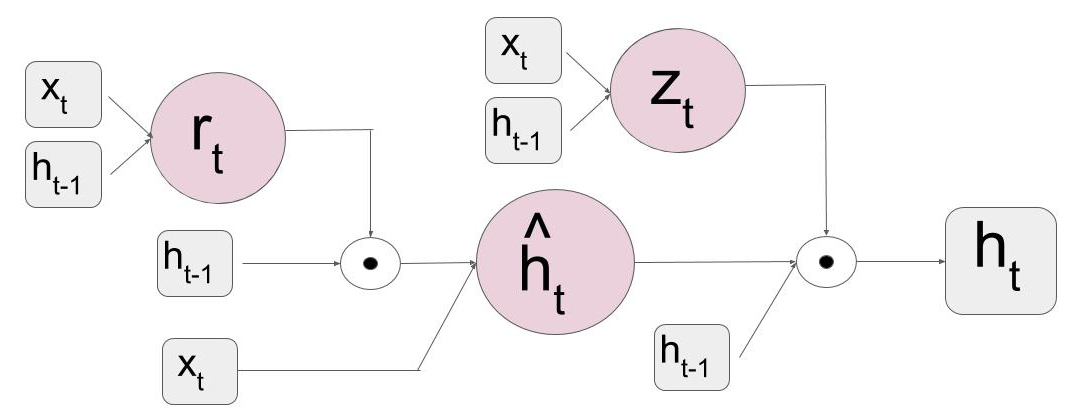}
    \caption{GRU Architecture.}
    \label{fig:gru}
\end{figure}

The Gated Recurrent Unit, similar to LSTM, utilizes a gated structure for flow-control. However, it has a simpler architecture which makes it both faster and less memory consuming. The model is shown in Figure \ref{fig:gru} and described in \ref{eq:gru} where $r_t$, $z_t$ is the reset and update gate respectively. While $h_t$ is the hidden state.
\begin{subequations}
\label{eq:gru}
\begin{align}
    z_t &= \sigma(W_{hz} h_{t-1} + W_{xz} x_t + b_z)\\
    r_t &= \sigma(W_{hr} h_{t-1} + W_{xr} x_t + b_r)\\
    \hat{h_t} &= \Phi(W_h (r_t \odot h_{t-1}) + W_x x_t + b)\\
    h_t &= (1-z_t)\odot h_{t-1} + z \odot \hat{h_t} 
\end{align}
\end{subequations}

GRU does not have direct control over memory content exposure while LSTM has it by having an output gate. These two are also different in the way that they update the memory nodes. LSTM updates its hidden state by summation over flow after input gate and forget gate. GRU however, assumes a correlation between how much to keep from the current state and how much to get from the previous state and it models this with the $z_t$ gate.

\section{Method} \label{method}
In an abstract view, we use a recurrent fully convolutional network (RFCN) that utilizes the temporal as well as spatial information for segmentation. The general theme in the design is to use recurrent nodes that combine fully convolutional network with a Recurrent unit(RU). The recurrent unit denotes either LSTM, GRU or Conv-GRU (which is explained in \ref{conv_gru}). In all of our networks, we aim for online segmentation in contrast to batch/offline version which needs the whole video as input. This is done by using a sliding window over the frames. Then, each window is propagated through the RFCN and yields a segmentation corresponding to the last frame in the sliding window. The recurrent layer can be employed to a sequence of feature maps or heat maps where each element in the series is the result of an image forward pass through an FCN network (Figure \ref{fig:overprop}). The whole network is trained end-to-end using pixel-wise classification logarithmic loss. 
We designed different network architectures to this end using conventional and also convolutional recurrent unit \ref{table:networks}. 

\subsection{Conventional Recurrent Unit for Segmentation} 
Our first architecture uses the Lenet network converted to a fully convolutional network as the base. Lenet is a well known and shallow network and as it is common, we used it for early experiments. We embed this model in recurrent node to allow the network to use temporal data. In Table \ref{table:networks}, RFC-Lenet is delineating this architecture. The output of deconvolution inside the fcn is a 2D map of dense predictions that is then flattened into 1D vector as input to the recurrent unit. The recurrent unit takes a vector from each frame in the sliding window and outputs the segmentation of the last frame (Figure \ref{fig:overprop}). \\
\newcolumntype{C}[1]{>{\centering\arraybackslash\hspace{0pt}}p{#1}}
%\newcolumntype{A}{>{\columncolor{blue!25}}c}

Note that utilizing the first architecture requires using large weight matrix in the RU layer since it operates on the flattened vectors of the full sized image. To lessen this problem, we can apply the deconvolution layer after the recurrent node. This will lead to our second architecture. The RU receives a vector of the flattened coarse map as its input and outputs the coarse predictions of the last frame in the sliding window. Then, deconvolution of this coarse map is used to get dense predictions. This proved to be useful with larger input images that lead to large number of parameters in the RU. Decreasing their size, allows the optimizer to search in a smaller state space and find a more generalized local minima in a faster training time. RFC-12s in the Table \ref{table:networks} is an example of this approach. This network is slightly modified version of the Lenet FCN. The main difference is that now the deconvolution comes as the last step after recursion being done. 

\subsection{Convolutional Gated Recurrent Unit (Conv-GRU) for Segmentation} \label{conv_gru}
\begin{figure*}[ht]
     \centering
    \includegraphics[width=\textwidth]{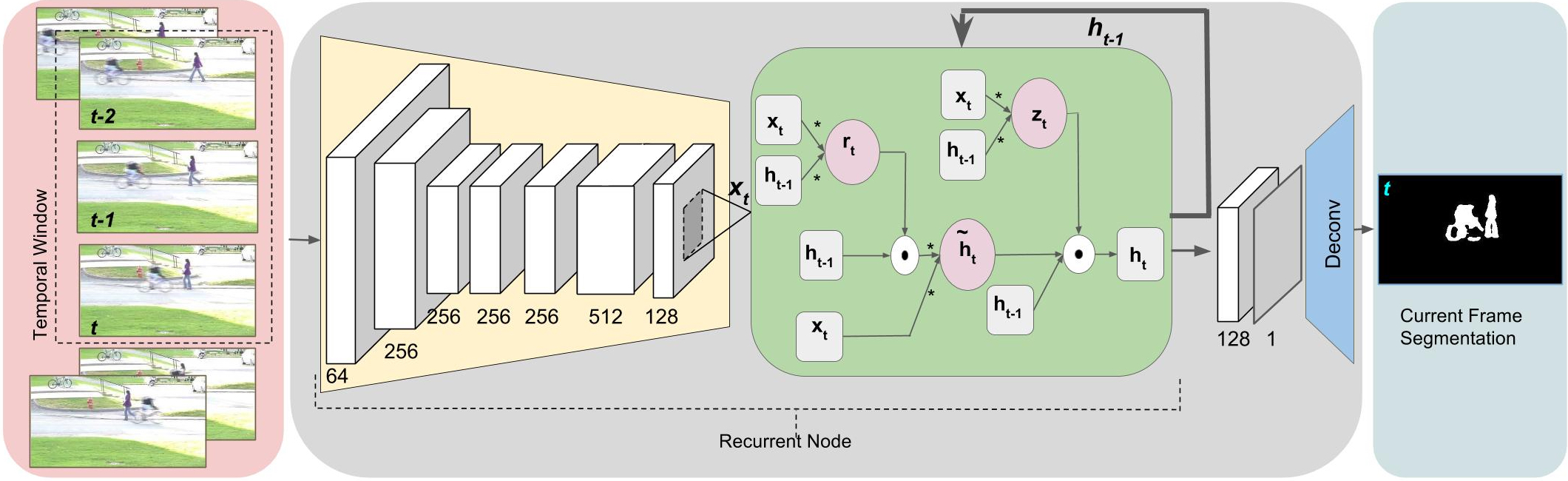}
    \caption{The architecture of RFC-VGG. Images are fed frame by frame into a recurrent FCN. A Conv-GRU layer is applied on the feature maps produced by the preceding network at each frame. The output of this layer goes to one more convolutional layer to generate heat maps. Finally, a deconvolution layer up-samples the heat map to the desired spatial size.}
    \label{fig:rfcvgg_detailed}
\end{figure*}

Conventional recurrent units are capable of processing temporal data however, their architecture is not suitable for working on images/feature maps for two reasons. 1) weights matrix size, 2) ignoring spatial connectivity. Assume a case where a recurrent unit is placed after a feature map with the spatial size of $h \times w$ and have a number of channels $c$. After flattening, it will turn into a $c \times h.w$ long matrix. Therefore, weights of the recurrent unit will be of size $c\times(h.w)^2$ which is power four of spatial dimension. These matrices for weights can only be maintained for small feature maps. Even if the computation was not an issue, such design introduces too much variance in the network which prevents generalization. 
In Convolutional recurrent units, similar to regular convolutional layer, weights are three dimensional and they convolve with the input instead of dot product. Accordingly, the cell's model, in the case of a GRU architecture, will turn into equations \ref{eq:conv_gru} where the dot products are replaced with convolutions. In this design, weights matrices are of size $k_h \times k_w \times c \times f$ where $k_h$, $k_w$, $c$ and $f$ are kernel's height, kernel's width, number of input channels, and number of filters, respectively. In Figure \ref{fig:gru} the operations applied on the input and the previous step will all be convolutions instead. Since we can assume spatial connectivity in feature maps, kernel size can be very small compared to feature map's spatial size. Therefore, this architecture is much more efficient and weights are easier to learn due to smaller search space.\\ 
We employ this approach for segmentation in a fully convolutional network. It is possible to apply this layer on either heat maps or feature maps. In the first case, the output of this layer will directly feed into the deconvolution layer and produces the pixel-wise probability map. In the latter case, at least one CNN layer needs to be used after the recurrent layer to convert its output feature maps to a heat map.\\
RFC-VGG in the Table \ref{table:networks} is an example of the second case. It is based on VGG-F \cite{simonyan2014very} network. Initializing weights of our filters by VGG-F trained weights, alleviates over-fitting problems as these weights are the result of extensive training on the imagenet. The network is cast to a fully convolutional one by replacing the fully connected layers with convolutional layers. The last two pooling layers are dropped from VGG-F to allow a finer segmentation. Then a convolutional gated recurrent unit is used followed by one convolutional layer and then deconvolution for up-sampling. Figure \ref{fig:rfcvgg_detailed} shows the detailed architecture of RFC-VGG.

\begin{subequations}
\label{eq:conv_gru}
\begin{align}
    z_t &= \sigma(W_{hz}\ast h_{t-1} + W_{xz}\ast x_t + b_z)\\
    r_t &= \sigma(W_{hr} \ast h_{t-1} + W_{xr} \ast x_t + b_r)\\
    \hat{h_t} &= \Phi(W_h \ast (r_t \odot h_{t-1}) + W_x \ast x_t + b)\\
    h_t &= (1-z_t)\odot h_{t-1} + z \odot \hat{h_t} 
\end{align}
\end{subequations} 

\section{Experiments} \label{experiments}
This section presents our experiments and results. First, we describe the datasets that we used then, we discuss our training methods and hyper-parameters settings. Finally, quantitative and qualitative results are shown.
\begin{figure}[ht]
\begin{subfigure}{.15\textwidth}
    \centering
    \includegraphics[scale= 0.5]{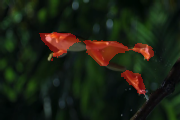}
\end{subfigure}%
\begin{subfigure}{.15\textwidth}
    \centering
    \includegraphics[scale= 0.5]{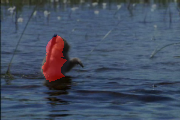}
\end{subfigure}%
\begin{subfigure}{.15\textwidth}
    \centering
    \includegraphics[scale= 0.5]{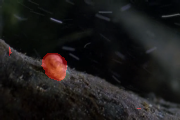}
\end{subfigure}%
\begin{subfigure}{.15\textwidth}
    \centering
    \includegraphics[scale= 0.5]{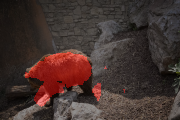}
\end{subfigure}%
\begin{subfigure}{.15\textwidth}
    \centering
    \includegraphics[scale= 0.5]{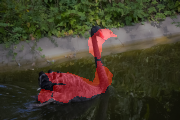}
\end{subfigure}%
\begin{subfigure}{.15\textwidth}
    \centering
    \includegraphics[scale= 0.5]{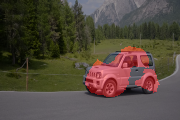}
\end{subfigure}

\begin{subfigure}{.15\textwidth}
    \centering
    \includegraphics[scale= 0.5]{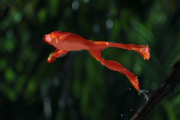}
\end{subfigure}%
\begin{subfigure}{.15\textwidth}
    \centering
    \includegraphics[scale= 0.5]{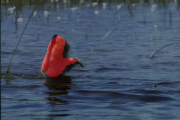}
\end{subfigure}%
\begin{subfigure}{.15\textwidth}
    \centering
    \includegraphics[scale= 0.5]{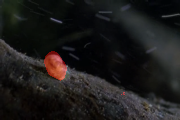}
\end{subfigure}%
\begin{subfigure}{.15\textwidth}
    \centering
    \includegraphics[scale= 0.5]{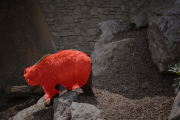}
\end{subfigure}%
\begin{subfigure}{.15\textwidth}
    \centering
    \includegraphics[scale= 0.5]{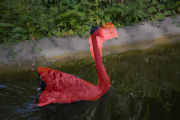}
\end{subfigure}%
\begin{subfigure}{.15\textwidth}
    \centering
    \includegraphics[scale= 0.5]{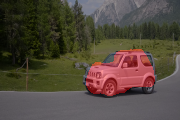}
\end{subfigure}
\caption{Qualtitative results over Segtrack V2 and Davis datasets, where top image has overlay of FC-VGG and bottom has RFC-VGG segmentation.}
\label{fig:qual}
\end{figure}

\definecolor{lightgray}{gray}{0.5}
\begin{table*}[ht!]
\centering
\caption{Details of proposed networks. $F(n)$ denotes filter size of $n\times n$. $P(n)$ denotes total of $n$ zero padding around the feature map. $S(n)$ denotes stride of length $n$ for the convolution. $D(n)$ denotes number of output feature maps from a particular layer $n$ for a layer (number of feature maps is same as previous layer if $D$ is not mentioned).}
\label{table:networks}
\begin{tabular}{|c|C{3cm}||c|C{3cm}||c|C{4cm}|}
\hline
 \multicolumn{6}{|c|}{Network Architectures}  \\ \hline \hline
 \multicolumn{2}{|c||}{RFC-Lenet} &\multicolumn{2}{|c||}{ RFC-12s}  & \multicolumn{2}{|c|}{RFC-VGG} \\ \hline
 \multicolumn{2}{|c||}{input: 28$\times$28 }& \multicolumn{2}{|c||}{input: 120$\times$180} &  \multicolumn{2}{|c|}{input: 240$\times$360}\\ \hline
 \multirow{17}{*}{\rotatebox[origin=c]{90}{Recurrent Node} } & Conv: F(5), P(10), D(20)& \multirow{16}{*}{\rotatebox[origin=c]{90}{Recurrent Node} }& Conv: F(5), S(3), P(10), D(20) & \multirow{15}{*}{\rotatebox[origin=c]{90}{Recurrent Node} }& Conv: F(11), S(4), P(40), D(64) \\ \cline{2-2} \cline{4-4} \cline{6-6}
 & Relu& & Relu  & & Relu  \\ \cline{2-2} \cline{4-4} \cline{6-6}
 & Pool 2$\times$2 &  & Pool 2$\times$2 & & Pool 3$\times$3  \\ \cline{2-2} \cline{4-4} \cline{6-6}
 & Conv: F(5), D(50)& &   Conv: F(5), D(50)&  & Conv: F(5), P(2) D(256)\\ \cline{2-2} \cline{4-4} \cline{6-6}
 & Relu& & Relu  & & Relu  \\ \cline{2-2} \cline{4-4} \cline{6-6}
 & Pool(2$\times$2)& &  Pool(2$\times$2)& & Pool(3$\times$3) \\ \cline{2-2} \cline{4-4} \cline{6-6}
 & Conv: F(3), D(500)& &Conv: F(3), D(500)  & & Conv: F(3), P(1) D(256)   \\ \cline{2-2} \cline{4-4} \cline{6-6}
 & Relu& & Relu  & & Relu  \\ \cline{2-2} \cline{4-4} \cline{6-6}
 & Conv: F(1), D(1)& & Conv: F(1), D(1) & & Conv: F(3), P(1) D(256) \\ \cline{2-2} \cline{4-4} \cline{6-6}
 & \multirow{ 5}{*}{-} & & \multirow{ 5}{*}{-} & & Relu  \\ \cline{6-6}
 & & & & &  Conv: F(3), P(1) D(256)\\ \cline{6-6}
 & & & & & Relu\\ \cline{6-6}
 & & & & & Conv: F(3), D(512) \\ \cline{6-6}
 & & & & & Conv: F(3), D(128) \\  \cline{2-2} \cline{4-4} \cline{6-6}
 & DeConv: F(10), S(4)& & Flatten & & ConvGRU: F(3), D(128) \\ \cline{2-2} \cline{4-4} \cline{5-6}
 & Flatten& & GRU: W(100$\times$100) &\cellcolor{lightgray} & Conv: F(1), D(1)\\ \cline{2-2} \cline{3-4} \cline{6-6}
 & GRU: W(784$\times$784) & \cellcolor{lightgray} &  DeConv: F(10), S(4) & \cellcolor{lightgray}& DeConv: F(20), S(8)\\ \hline
 
\end{tabular}
\end{table*}

All the experiments are performed on our own implemented library. To our knowledge, there is no open source framework that accommodates RFCNN architecture. Therefore, we built our own library on top of Theano \cite{Bastien-Theano-2012} to create arbitrary networks that have can have FCN as a recursive module. Key features of this implementation are: \textbf{(1)} Supports networks with the temporal operation for images. The architecture can be any arbitrary CNN and an arbitrary number of recurrent layers. The network supports any length input. \textbf{(2)} Three gated architecture, LSTM, GRU, and Conv-GRU is available for the recurrent layer. \textbf{(3)} Deconvolution layer and skip architecture to support segmentation with FCN.

\subsection{Datasets}
In this paper four datasets are used: 1) Moving MNIST. 2) Change detection\cite{goyette2012changedetection}. 3) Segtrack version 2\cite{li2013video}. 4) Densely Annotated VIdeo Segmentation (Davis) \cite{perazzibenchmark}. Figure \ref{fig:qual} shows samples from the latter two.

\textbf{Moving MNIST} dataset is synthesized from original MNIST by moving the characters in random but consistent directions. The labels for segmentation is generated by thresholding input images after translation. We consider each translated image as a new frame. Therefore we can have arbitrary length sequence of images. 

\textbf{Change Detection Dataset\cite{goyette2012changedetection}} This dataset provides realistic, diverse set of videos with pixel-wise labeling of moving objects. The dataset includes both indoor and outdoor scenes. It focuses on moving object segmentation. In the motion detection, we were looking for videos that have similar moving objects, e.g. cars or humans so that there would be semantic correspondences among sequences. Accordingly, we chose six videos: Pedestrians, PETS2006, Badminton, CopyMachine, Office, and Sofa. 

\textbf{SegTrack V2\cite{li2013video}} is a collection of fourteen video sequences with objects of interest manually segmented. The dataset has sequences with both single or multiple objects. In the latter case, we consider all the segmented objects as one and we perform foreground segmentation.

\textbf{Davis\cite{perazzibenchmark}} dataset has fifty high resolution and densely annotated videos with pixel accurate groundtruth. The videos include multiple challenges such as occlusions, fast motion, nonlinear deformation and motion blur. 

\subsection{Results}
The main experiments are conducted using Adadelta \cite{zeiler2012adadelta} for optimization that practically gave much faster convergence than standard stochastic gradient descent. The logistic loss function is used and the maximum number of epochs used for the training is 500. The evaluation metrics used are precision, recall, F-measure and IoU, shown in equation \ref{eq-prec} and \ref{eq-fmes} where tp, fp, fn denote true positives, false positives, and false negatives respectively. 

\begin{equation}
    precision= \frac{tp}{tp+fp}, 
    recall= \frac{tp}{tp+fn}
    \label{eq-prec}
\end{equation}

\begin{equation}
    F-measure= \frac{2*precision*recall}{precision+recall}
    \label{eq-fmes}
\end{equation}
\begin{equation}
    IoU = \frac{tp}{tp+fp+fn}
    \label{eq-fmes}
\end{equation}

\newcolumntype{L}[1]{>{\raggedright\arraybackslash\hspace{0pt}}p{#1}}
{\setlength{\extrarowheight}{10pt}
\begin{table*}[ht!]
\centering
\caption{Comparison of RFC-VGG with its baseline counterpart on DAVIS and SegTrack}
\label{table:segtrackdavis}
\begin{tabular}{|l|L{3cm}|c|c|c|c|}
\hline
\multicolumn{2}{|l|}{} & Precision & Recall & F-measure & IoU \\ \hline
\multirow{2}{*}{ SegTrack V2} &  FC-VGG & 0.7759 & 0.6810 & 0.7254 & 0.7646 \\ \cline{2-6} 
                   & RFC-VGG & \textbf{0.8325} & \textbf{0.7280} & \textbf{0.7767} & \textbf{0.8012} \\ \hline
\multirow{2}{*}{ DAVIS} & FC-VGG & 0.6834 & 0.5454 & 0.6066 & 0.6836 \\ \cline{2-6} 
                   & RFC-VGG & \textbf{0.7233} & \textbf{0.5586} & \textbf{0.6304} & \textbf{0.6984} \\ \hline
\end{tabular}
\end{table*}
In this set of experiments, a fully convolutional VGG is used as a baseline denoted as FC-VGG. It is compared against the recurrent version RFC-VGG. The initial five convolutional layers are not finetuned and are from the pretrained VGG architecture to avoid overfitting the data. The results of the experiments on SegTrackV2 and Densely Annotated VIdeo Segmentation (DAVIS) datasets are provided in \ref{table:segtrackdavis}. In these experiments, the data of each sequence is divided into two splits with half as training data and the other half as keep out test data. It is apparent in the results that RFC-VGG outperforms the FC-VGG architecture on both datasets with about 3\% and 5\% on DAVIS and SegTrack respectively.\\

\par Figure \ref{fig:qual} shows the qualitative analysis of RFC-VGG against FC-VGG. It shows that utilizing temporal information through the recurrent unit gives better segmentation for the object. This can be explained due to the implicit learning of the motion that segmented objects undergo in the recurrent units. It also shows that using conv-GRU as the recurrent unit can enable the extraction of temporal information from feature maps due to the reduced parameter set. Thus the recurrent unit can learn the motion pattern of the segmented objects by working on richer information from these feature maps. It's also worth noting that the performance of the RFCN network depends on its baseline fully convolutional counter part. Thus with an enhanced fully convolutional network using skip architecture, to get finer segmentation, the recurrent version should improve as well.

\subsection{Further Analysis}
In this section, we present our experiments using conventional recurrent layers for segmentation. These experiments provide further analysis on different recurrent units. They also compare end-to-end architecture versus the decoupled one. We use moving MNIST and change detection datasets for this part.\\
Images of MNIST dataset are relatively small (28$\times$28) which allows us to test our RFC-Lenet(a) network \ref{table:networks}. A fully convolutional Lenet is compared against RFC-Lenet(a). Table \ref{table_mnist} shows the results that were obtained. The results of RFC-Lenet with GRU is better than FC-Lenet with 2\% improvement. But segmentation of MNIST characters is an easy task as it ends up to learning thresholding. Note also that GRU gave better results than LSTM.

\begin{table}[!htb]
  \centering
  \caption{Precision, Recall, and F-measure on FC-Lenet, LSTM, GRU, and RFC-Lenet tested on synthesized MNIST dataset}
  \begin{tabular}{| L{2cm} | C{1.5cm} | C{1.5cm} | C{1.5cm} |}
    \hline
     & Precision & Recall & F-measure\\ \hline
    FC-Lenet & 0.868 & \textbf{0.922} & 0.894\\ \hline
    LSTM & 0.941 & 0.786 & 0.856 \\ \hline
    GRU & 0.955 & 0.877 & 0.914 \\ \hline
    RFC-Lenet & \textbf{0.96} & 0.877 & \textbf{0.916}\\
    \hline
  \end{tabular}
 
 \label{table_mnist}
\end{table}

The second set of experiments is conducted on real data from the motion detection benchmark using RFC-12s. Throughout these experiments, the training constitutes 70\% of the sequences and 30\% for the test set. The experiments compared the recurrent convolutional network trained in an end-to-end fashion denoted as RFC-12s with its baseline, the fully convolutional network FC-12s.
It is also compared against the decoupled training of the FC-12s and the recurrent unit. Where GRU is trained on the probability maps output from FC-12s. Table \ref{table_alldata} shows the results of these experiments, where the RFC-12s network had a 1.4\% improvement over FC-12s. We observe less relative improvement compared to using Conv-GRU because in regular GRU spatial connectivities are ignored. However, incorporating the temporal data still helped the segmentation accuracy.

\begin{table}[!htb]
  \centering
   \caption{Precision, Recall, and F-measure on architectures FCN-12s, GRU pretraining on coarse map from FCN-12s, RFC-12s on six sequences from motion detection benchmark on the test set. (D) and (EE) indicate the decoupled and the end-to-end integration of recurrent units with the FCN, respectively. }
  \begin{tabular}{| L{2.5cm} | C{1.4cm} | C{1cm} | C{1.5cm} |}
   \hline
     & Precision & Recall & F-measure\\ \hline
    FC-12s & 0.827 & 0.585 & 0.685 \\ \hline
    RFC-12s (D) & \textbf{0.835} & 0.587 & 0.69\\ \hline
    RFC-12s (EE) & 0.797 & \textbf{0.623} & \textbf{0.7}\\
    \hline
  \end{tabular}
   \label{table_alldata}
\end{table}

\section{Conclusion and Future Work} \label{conclusion}
In this paper, we presented a novel approach in incorporating temporal information for video segmentation. This approach utilizes recurrent units on top of a fully convolutional network as a mean to include previously seen frames to decide the segmentation for the current frame. The paper also proposes using convolutional recurrent units (conv-gru in particular) for segmentation. We tested the method on both synthesized and real data for the segmentation task for different architectures. We showed that by having a recurrent layer after either probability map or feature map can improve the performance of the segmentation. The proposed architecture can process arbitrary length videos which makes it suitable for both batch and online scenarios.\\

For the future work, We want to extend our framework to semantic video segmentation using multiple recurrent layers for richer dynamic representation. We also want to apply our method on newer networks that are proposed for per image segmentation as in \cite{visin2015reseg}.

\bibliographystyle{plain}
\bibliography{main}

\begin{thebibliography}{10}

\bibitem{ballas2015delving}
Nicolas Ballas, Li~Yao, Chris Pal, and Aaron Courville.
\newblock Delving deeper into convolutional networks for learning video
  representations.
\newblock {\em arXiv preprint arXiv:1511.06432}, 2015.

\bibitem{Bastien-Theano-2012}
Fr{\'{e}}d{\'{e}}ric Bastien, Pascal Lamblin, Razvan Pascanu, James Bergstra,
  Ian~J. Goodfellow, Arnaud Bergeron, Nicolas Bouchard, and Yoshua Bengio.
\newblock Theano: new features and speed improvements.
\newblock Deep Learning and Unsupervised Feature Learning NIPS 2012 Workshop,
  2012.

\bibitem{bengio1994learning}
Yoshua Bengio, Patrice Simard, and Paolo Frasconi.
\newblock Learning long-term dependencies with gradient descent is difficult.
\newblock {\em Neural Networks, IEEE Transactions on}, 5(2):157--166, 1994.

\bibitem{cho2014properties}
Kyunghyun Cho, Bart van Merri{\"e}nboer, Dzmitry Bahdanau, and Yoshua Bengio.
\newblock On the properties of neural machine translation: Encoder-decoder
  approaches.
\newblock {\em arXiv preprint arXiv:1409.1259}, 2014.

\bibitem{chung2014empirical}
Junyoung Chung, Caglar Gulcehre, KyungHyun Cho, and Yoshua Bengio.
\newblock Empirical evaluation of gated recurrent neural networks on sequence
  modeling.
\newblock {\em arXiv preprint arXiv:1412.3555}, 2014.

\bibitem{donahue2015long}
Jeffrey Donahue, Lisa Anne~Hendricks, Sergio Guadarrama, Marcus Rohrbach,
  Subhashini Venugopalan, Kate Saenko, and Trevor Darrell.
\newblock Long-term recurrent convolutional networks for visual recognition and
  description.
\newblock In {\em Proceedings of the IEEE Conference on Computer Vision and
  Pattern Recognition}, pages 2625--2634, 2015.

\bibitem{goyette2012changedetection}
Nil Goyette, Pierre-Marc Jodoin, Fatih Porikli, Janusz Konrad, and Prakash
  Ishwar.
\newblock Changedetection. net: A new change detection benchmark dataset.
\newblock In {\em Computer Vision and Pattern Recognition Workshops (CVPRW),
  2012 IEEE Computer Society Conference on}, pages 1--8. IEEE, 2012.

\bibitem{graves2013generating}
Alex Graves.
\newblock Generating sequences with recurrent neural networks.
\newblock {\em arXiv preprint arXiv:1308.0850}, 2013.

\bibitem{hochreiter1997long}
Sepp Hochreiter and J{\"u}rgen Schmidhuber.
\newblock Long short-term memory.
\newblock {\em Neural computation}, 9(8):1735--1780, 1997.

\bibitem{huang2015densebox}
Lichao Huang, Yi~Yang, Yafeng Deng, and Yinan Yu.
\newblock Densebox: Unifying landmark localization with end to end object
  detection.
\newblock {\em arXiv preprint arXiv:1509.04874}, 2015.

\bibitem{johnson2015densecap}
Justin Johnson, Andrej Karpathy, and Li~Fei-Fei.
\newblock Densecap: Fully convolutional localization networks for dense
  captioning.
\newblock {\em arXiv preprint arXiv:1511.07571}, 2015.

\bibitem{karpathy2014large}
Andrej Karpathy, George Toderici, Sanketh Shetty, Thomas Leung, Rahul
  Sukthankar, and Li~Fei-Fei.
\newblock Large-scale video classification with convolutional neural networks.
\newblock In {\em Proceedings of the IEEE conference on Computer Vision and
  Pattern Recognition}, pages 1725--1732, 2014.

\bibitem{krizhevsky2012imagenet}
Alex Krizhevsky, Ilya Sutskever, and Geoffrey~E Hinton.
\newblock Imagenet classification with deep convolutional neural networks.
\newblock In {\em Advances in neural information processing systems}, pages
  1097--1105, 2012.

\bibitem{li2013video}
Fuxin Li, Taeyoung Kim, Ahmad Humayun, David Tsai, and James~M Rehg.
\newblock Video segmentation by tracking many figure-ground segments.
\newblock In {\em Proceedings of the IEEE International Conference on Computer
  Vision}, pages 2192--2199, 2013.

\bibitem{lin2015efficient}
Guosheng Lin, Chunhua Shen, Ian Reid, et~al.
\newblock Efficient piecewise training of deep structured models for semantic
  segmentation.
\newblock {\em arXiv preprint arXiv:1504.01013}, 2015.

\bibitem{long2015fully}
Jonathan Long, Evan Shelhamer, and Trevor Darrell.
\newblock Fully convolutional networks for semantic segmentation.
\newblock In {\em Proceedings of the IEEE Conference on Computer Vision and
  Pattern Recognition}, pages 3431--3440, 2015.

\bibitem{michalski2014modeling}
Vincent Michalski, Roland Memisevic, and Kishore Konda.
\newblock Modeling deep temporal dependencies with recurrent grammar cells"".
\newblock In {\em Advances in neural information processing systems}, pages
  1925--1933, 2014.

\bibitem{noh2015learning}
Hyeonwoo Noh, Seunghoon Hong, and Bohyung Han.
\newblock Learning deconvolution network for semantic segmentation.
\newblock In {\em Proceedings of the IEEE International Conference on Computer
  Vision}, pages 1520--1528, 2015.

\bibitem{pavel2015recurrent}
Mircea~Serban Pavel, Hannes Schulz, and Sven Behnke.
\newblock Recurrent convolutional neural networks for object-class segmentation
  of rgb-d video.
\newblock In {\em Neural Networks (IJCNN), 2015 International Joint Conference
  on}, pages 1--8. IEEE, 2015.

\bibitem{perazzibenchmark}
F~Perazzi, J~Pont-Tuset1~B McWilliams, L~Van~Gool, M~Gross, and
  A~Sorkine-Hornung.
\newblock A benchmark dataset and evaluation methodology for video object
  segmentation.

\bibitem{simonyan2014very}
Karen Simonyan and Andrew Zisserman.
\newblock Very deep convolutional networks for large-scale image recognition.
\newblock {\em arXiv preprint arXiv:1409.1556}, 2014.

\bibitem{sutskever2011generating}
Ilya Sutskever, James Martens, and Geoffrey~E Hinton.
\newblock Generating text with recurrent neural networks.
\newblock In {\em Proceedings of the 28th International Conference on Machine
  Learning (ICML-11)}, pages 1017--1024, 2011.

\bibitem{szegedy2015going}
Christian Szegedy, Wei Liu, Yangqing Jia, Pierre Sermanet, Scott Reed, Dragomir
  Anguelov, Dumitru Erhan, Vincent Vanhoucke, and Andrew Rabinovich.
\newblock Going deeper with convolutions.
\newblock In {\em Proceedings of the IEEE Conference on Computer Vision and
  Pattern Recognition}, pages 1--9, 2015.

\bibitem{taylor2010convolutional}
Graham~W Taylor, Rob Fergus, Yann LeCun, and Christoph Bregler.
\newblock Convolutional learning of spatio-temporal features.
\newblock In {\em Computer Vision--ECCV 2010}, pages 140--153. Springer, 2010.

\bibitem{vinyals2012revisiting}
Oriol Vinyals, Suman~V Ravuri, and Daniel Povey.
\newblock Revisiting recurrent neural networks for robust asr.
\newblock In {\em Acoustics, Speech and Signal Processing (ICASSP), 2012 IEEE
  International Conference on}, pages 4085--4088. IEEE, 2012.

\bibitem{visin2015reseg}
Francesco Visin, Kyle Kastner, Aaron Courville, Yoshua Bengio, Matteo
  Matteucci, and Kyunghyun Cho.
\newblock Reseg: A recurrent neural network for object segmentation.
\newblock {\em arXiv preprint arXiv:1511.07053}, 2015.

\bibitem{wang2015visual}
Lijun Wang, Wanli Ouyang, Xiaogang Wang, and Huchuan Lu.
\newblock Visual tracking with fully convolutional networks.
\newblock In {\em Proceedings of the IEEE International Conference on Computer
  Vision}, pages 3119--3127, 2015.

\bibitem{zeiler2012adadelta}
Matthew~D Zeiler.
\newblock Adadelta: an adaptive learning rate method.
\newblock {\em arXiv preprint arXiv:1212.5701}, 2012.

\bibitem{zheng2015conditional}
Shuai Zheng, Sadeep Jayasumana, Bernardino Romera-Paredes, Vibhav Vineet,
  Zhizhong Su, Dalong Du, Chang Huang, and Philip~HS Torr.
\newblock Conditional random fields as recurrent neural networks.
\newblock In {\em Proceedings of the IEEE International Conference on Computer
  Vision}, pages 1529--1537, 2015.

\end{thebibliography}
\end{document}